\documentclass{article}

\usepackage{microtype}
\usepackage{graphicx}
\usepackage{hyperref}

\usepackage[preprint]{icml2026} 
\usepackage[table,dvipsnames]{xcolor}  
\usepackage{amsmath}
\usepackage{amssymb}
\usepackage{mathtools}
\usepackage{amsthm}
\usepackage{titlesec}
\titlespacing*{\section}{0pt}{4pt}{0pt}
\titlespacing*{\subsection}{0pt}{3pt}{2pt}
\titlespacing*{\subsubsection}{0pt}{0pt}{0pt}
\usepackage{caption}
\usepackage{subcaption}
\usepackage{graphicx}
\usepackage{booktabs}
\usepackage{multirow}
\usepackage{arydshln}
\makeatletter
\def\adl@colcolor{}   
\makeatother
\definecolor{lightgray}{gray}{0.92}
\usepackage[capitalize,noabbrev]{cleveref}

\theoremstyle{plain}

\theoremstyle{definition}

\theoremstyle{remark}

\usepackage[textsize=tiny]{todonotes}
\usepackage{enumitem}
\setlist{nosep}
\icmltitlerunning{Consensus-Guided Test-Time Learning}

\begin{document}
\setlength{\parskip}{0.3ex}

\twocolumn[
  \icmltitle{DiSCTT: Consensus-Guided Self-Curriculum for Efficient Test-Time Adaptation in Reasoning
  }




  \icmlsetsymbol{equal}{*}

  \begin{icmlauthorlist}
    \icmlauthor{Mohammad Mahdi Moradi}{yyy}
    \icmlauthor{Sudhir Mudur}{yyy}
  \end{icmlauthorlist}

  \icmlaffiliation{yyy}{Department of Compute Science, Concordia University, Montreal, Canada}

  \icmlcorrespondingauthor{Sudhir Mudur}{mudur@cse.concordia.ca}

  \icmlkeywords{Machine Learning, ICML}

  \vskip 0.3in
]



\printAffiliationsAndNotice{}  

\begin{abstract}
Test-time adaptation offers a promising avenue for improving reasoning performance in large language models without additional supervision, but existing approaches often apply a uniform optimization objective across all inputs, leading to inefficient or unstable adaptation on heterogeneous reasoning problems. We propose DiSCTT, a difficulty-aware, consensus-guided self-curriculum framework that dynamically allocates test-time optimization strategies based on instance-level epistemic uncertainty estimated from agreement among sampled reasoning trajectories. Inputs with high consensus are consolidated via supervised fine-tuning using majority-agreed solutions as pseudo-labels, while low-consensus inputs are optimized via reinforcement learning with a consensus-regularized objective that encourages diversity under relevance constraints. Across a broad suite of mathematical and general reasoning benchmarks, DiSCTT consistently outperforms strong test-time adaptation baselines, achieving higher accuracy with reduced variance and substantially lower computation and wall-clock training times. These results demonstrate that explicitly accounting for instance difficulty and uncertainty enables more stable, efficient, and effective test-time adaptation for reasoning models.

\end{abstract}

\section{Introduction}

Large language models (LLMs) have achieved strong performance on mathematical and general reasoning tasks through supervised fine-tuning, chain-of-thought prompting, and reinforcement learning \cite{wei2022cot}. Despite these advances, inference-time behavior remains largely static: once deployed, a model applies a fixed policy to all inputs, regardless of input difficulty or the model’s own uncertainty. This rigidity has motivated growing interest in \emph{test-time adaptation}, where models update their behavior during inference without access to ground-truth labels \cite{goyal2022revisiting, zhang2023testtime}. Reasoning tasks are a particularly natural setting for such adaptation, as they exhibit wide variation in difficulty and admit multiple valid solution paths.

\begin{figure}[h]
    \centering
    \includegraphics[scale=0.4]{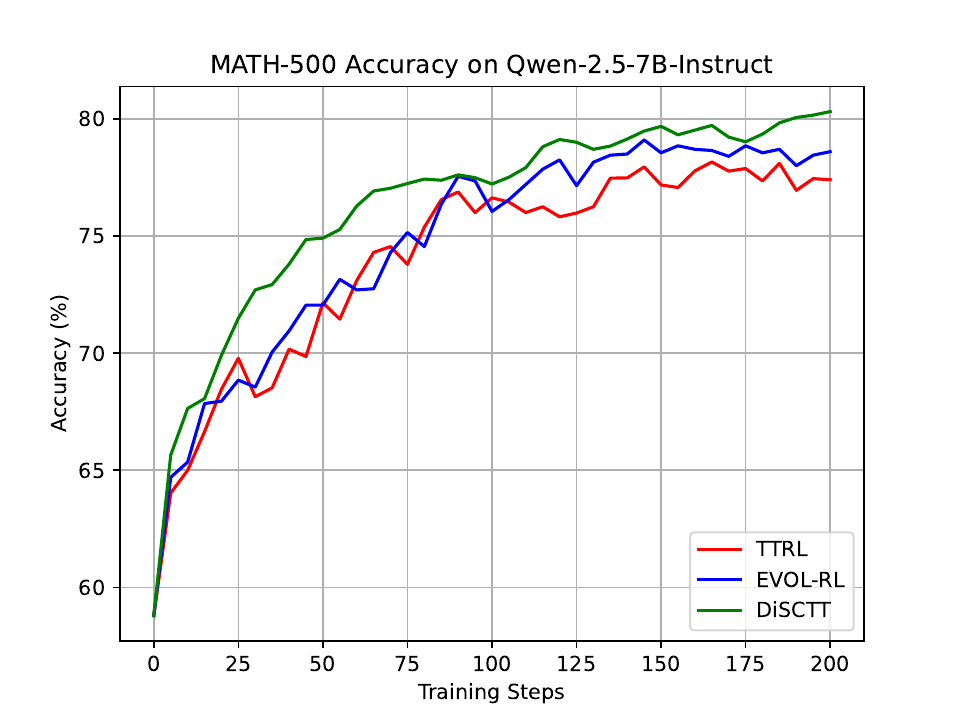}
\caption{
Accuracy on MATH-500 over test-time training updates for different adaptation strategies. Uniform test-time training baselines show early saturation or unstable gains, whereas DiSCTT achieves more stable and sustained improvements through difficulty-aware routing.
    }
    \label{fig:1}
    \vspace{-6pt}
\end{figure}

\begin{figure}[h]
    \centering
    \includegraphics[scale=0.4]{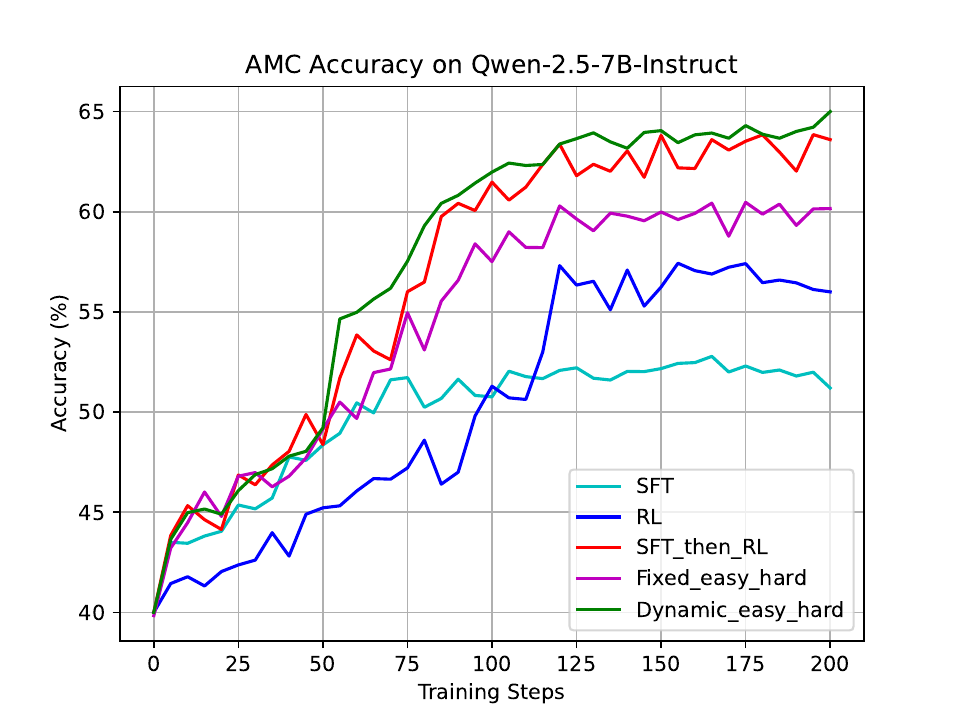}
    \caption{
    AMC accuracy for different adaptation strategies.  Difficulty-aware routing and correctness-gated exploration improve convergence stability and prevent performance collapse observed under uniform RL.
    }
    \label{fig:2}
    \vspace{-18pt}
\end{figure}

A key challenge in test-time adaptation is that reasoning problems are inherently heterogeneous. Easier instances often benefit from consolidation, where reinforcing high-confidence solutions stabilizes correct behavior. Harder instances, by contrast, require structured exploration to uncover alternative reasoning strategies. Most existing test-time training methods apply a single optimization objective uniformly across all inputs (Figures~\ref{fig:1},\ref{fig:2}). Supervised self-training can saturate quickly and offers limited benefit on difficult problems. On the other hand, uniform reinforcement learning introduces unnecessary variance on already-solved instances, often leading to unstable or inefficient learning dynamics.

These limitations point to a broader principle: effective test-time adaptation for reasoning requires allocating different learning objectives to different inputs based on model uncertainty. In particular, test-time learners must (i) estimate instance difficulty online, (ii) consolidate high-confidence solutions efficiently, and (iii) focus exploration on uncertain instances while stabilizing reinforcement learning in the absence of external supervision. Applying a single objective uniformly at inference time inevitably trades off over-optimizing easy instances against under-exploring genuinely hard ones.

Reasoning problems also pose challenges for uncertainty estimation. Token-level confidence scores or entropy are poorly aligned with multi-step reasoning, where errors often emerge only at the trajectory level \cite{geng2024survey}. In contrast, agreement among independently sampled reasoning trajectories can provide a useful proxy for epistemic uncertainty: high consensus indicates stable reasoning modes, while disagreement reflects uncertainty over inference strategies. This trajectory-level consensus therefore serves as an effective, label-free estimate of relative instance difficulty at test time.

We operationalize this principle with DiSCTT (Difficulty-aware Consensus-Guided Self-Curriculum Test-Time Adaptation). DiSCTT uses consensus among sampled reasoning completions to estimate epistemic uncertainty online and dynamically routes inputs between two complementary learning objectives. Inputs with high consensus are optimized via supervised fine-tuning using majority-agreed solutions as pseudo-labels, consolidating correct behavior with low variance. Inputs with low consensus are optimized via reinforcement learning, enabling structured exploration of alternative reasoning paths. This routing is periodically recomputed, yielding a self-evolving curriculum that adapts to the model’s changing competence rather than relying on a fixed easy--hard decomposition.

We summarize our contributions as follows:
\begin{itemize}
\item \textbf{Consensus-based difficulty estimation at test time.}
We formalize agreement among independently sampled reasoning trajectories as an online, instance-level estimator of epistemic uncertainty, enabling difficulty-aware adaptation without access to ground-truth labels.

\item \textbf{Difficulty-aware self-curriculum for test-time adaptation.}
We introduce a dynamic routing mechanism that allocates supervised consolidation to high-consensus inputs and reinforcement learning to low-consensus inputs, yielding a self-evolving curriculum that adapts to the model’s changing competence during inference.

\item \textbf{Stabilized label-free reinforcement learning for reasoning.}
We propose a correctness-gated, population-relative reward augmented with relevance-aware semantic gating, which encourages controlled exploration over reasoning trajectories while improving the stability of policy updates in the absence of external supervision.

\item \textbf{Extensive empirical evaluation across models and benchmarks.}
We show that DiSCTT consistently improves accuracy, stability, and computational efficiency across diverse mathematical and general reasoning benchmarks, model scales, and test-time budgets.
\end{itemize}


\section{Related Work}

Modern reasoning performance in large language models is closely tied to eliciting intermediate reasoning steps and aggregating multiple sampled solutions. Chain-of-thought prompting improves multi-step reasoning by inducing explicit intermediate steps \cite{wei2022chain}. Beyond single-sample decoding, self-consistency demonstrates that sampling many reasoning paths and selecting the most frequent final answer via majority voting yields substantial gains across reasoning tasks \cite{wang2022self}. This use of agreement among model generations as a weak supervision signal underpins a growing class of label-free or weakly supervised approaches for improving reasoning without access to ground-truth labels at inference  time. However, in most prior works, consensus is used as a static selection mechanism rather than as a dynamic signal to guide how learning objectives should be allocated.

Test-Time Training (TTT) formalizes adaptation at inference by updating model parameters on unlabeled test inputs using self-supervised objectives, primarily to address distribution shift \cite{sun2020test}. Verifier-Driven Selective Test-Time Training (VDS-TTT) extends this paradigm by conditioning supervised fine-tuning updates using an external verifier \cite{moradi2025continuous}. While verifier-based selection can improve stability, performance becomes strongly dependent on verifier quality and availability, and such methods do not model population-level difficulty or allocate exploration across reasoning instances.

In reasoning-specific settings, Test-Time Reinforcement Learning (TTRL) extends TTT by applying reinforcement learning updates at inference time using majority-based signals derived from multiple generations \cite{zuo2025ttrl}. TTRL establishes that label-free self-improvement is feasible for complex reasoning tasks, but also exposes two key challenges. First, updates are applied uniformly, remaining blind to epistemic uncertainty and problem difficulty. Second, repeated self-updates under uniform reinforcement pressure can introduce instability, noisy learning dynamics, or partial collapse.

To mitigate exploration collapse in label-free reinforcement learning, EVOL-RL proposes coupling majority-based selection with novelty-promoting variation, implemented via embedding-space diversity and GRPO-based optimization \cite{zhou2025evolving}. EVOL-RL applies a uniform optimization objective across all inputs and while effective, does not explicitly account for heterogeneity in instance difficulty. Consequently, novelty-driven exploration is encouraged even for high-confidence, already-solved instances, introducing unnecessary variance and potentially destabilizing learning. In addition, novelty is defined globally over the response population, without conditioning on instance-specific reasoning structure or semantic relevance, which can promote spurious or off-task deviations.

Orthogonal to how test-time learning signals are constructed, reinforcement learning has become a standard optimization backbone for improving reasoning and alignment in language models, including RLHF-style pipelines \cite{christiano2017deep, ouyang2022training} and reinforcement learning from verifiable rewards in mathematical reasoning. Group Relative Policy Optimization (GRPO), introduced in DeepSeekMath, has emerged as a practical optimizer for reasoning-centric reinforcement learning due to its stability and efficiency under group-based sampling \cite{shao2024deepseekmath}. In this work, GRPO is adopted purely as an optimization backbone, while our contributions focus on when reinforcement learning should be applied and how exploration should be stabilized under fully label-free conditions.

Prior work establishes three complementary directions: label-free test-time adaptation (TTRL), selective or verifier-driven updates (VDS-TTT), and diversity-promoting reinforcement learning without external supervision (EVOL-RL). DiSCTT unifies and extends these paradigms through a difficulty-aware allocation of learning objectives, rather than applying reinforcement learning uniformly or modulating rewards within a single objective. Specifically, DiSCTT addresses difficulty blindness via consensus-based epistemic routing, stabilizes optimization through correctness-gated reinforcement learning with relevance-aware semantic gating, and replaces global semantic novelty with population-aware distributional divergence over majority-correct reasoning trajectories. Together, these choices enable stable, scalable, and fully label-free test-time self-improvement for reasoning-centric language models. More broadly, DiSCTT is also related to prior work on adaptive computation, curriculum learning, and uncertainty-aware optimization, which allocate computation or learning effort based on estimates of confidence or difficulty.

\begin{figure*}[ht]
  \vskip 0.2in
  \begin{center}
    \centerline{\includegraphics[scale=0.5]{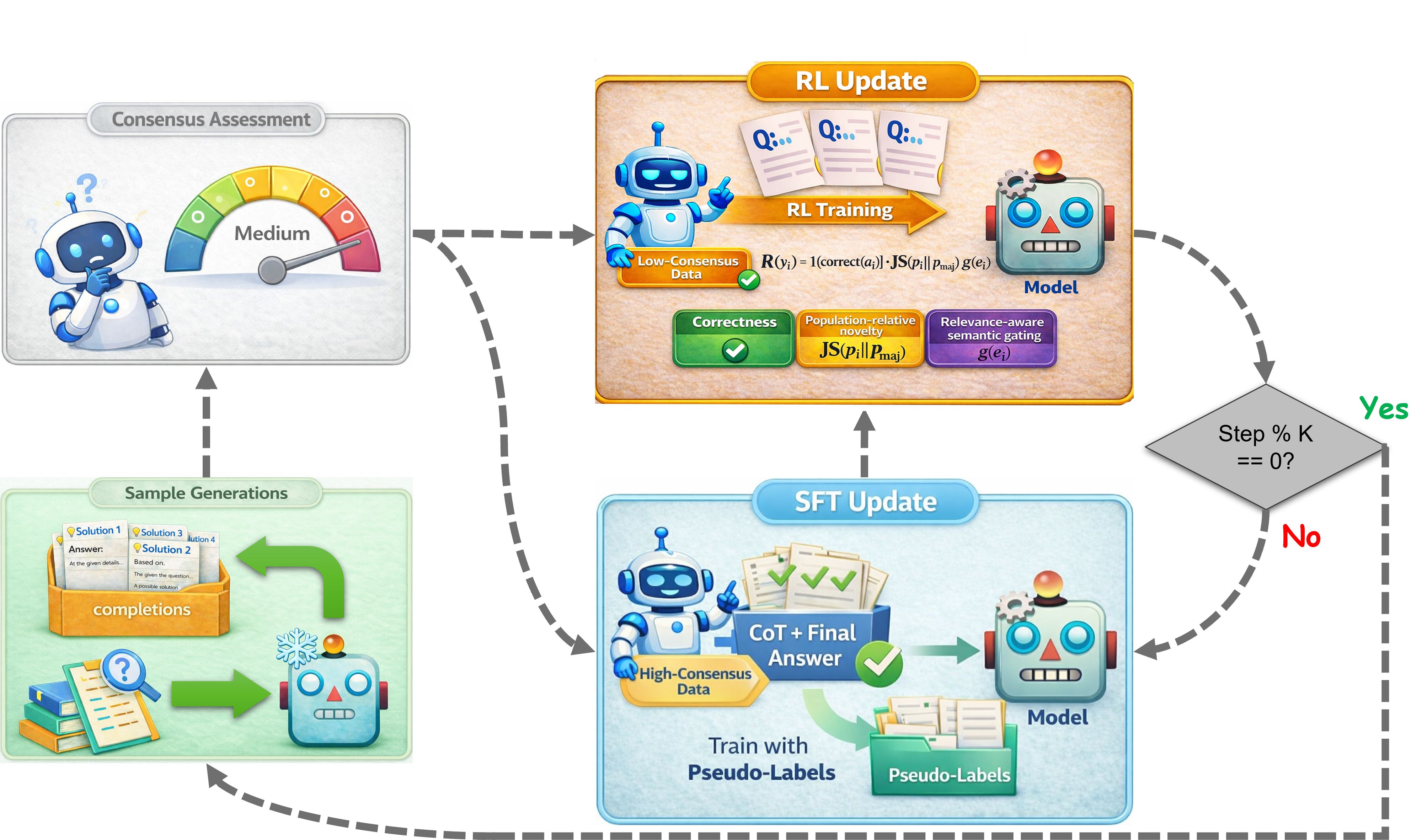}}
\caption{
Overview of DiSCTT. For each input, multiple reasoning completions are sampled and evaluated for consensus. High-consensus inputs are routed to supervised fine-tuning using majority-agreed solutions as pseudo-labels, while low-consensus inputs are optimized via reinforcement learning. This difficulty-aware bifurcation is periodically recomputed, yielding a self-evolving test-time curriculum.}
    \label{fig:framework}
  \end{center}
 \vspace{-24pt}
\end{figure*}

\section{Method}
We present DiSCTT, a test-time adaptation framework that allocates learning objectives based on instance difficulty. For each input, the model samples multiple reasoning completions and uses their agreement to estimate epistemic uncertainty, which governs routing between supervised fine-tuning and reinforcement learning. High-consensus inputs are consolidated via supervised fine-tuning, while low-consensus inputs are optimized via reinforcement learning, avoiding unnecessary high-variance updates on already-solved instances. Figure~\ref{fig:framework} provides an overview of the DiSCTT framework, illustrating consensus-based difficulty estimation, dynamic routing between learning objectives, and the resulting self-evolving curriculum. Reinforcement learning is applied only to a dynamically selected subset of difficult inputs, while high-confidence instances are consolidated through supervised fine-tuning, enabling stable and efficient test-time reasoning improvement.
We next describe how consensus induces a difficulty-aware partition of test-time inputs, followed by the supervised and reinforcement learning updates applied to each partition.

\subsection{Problem Setting}
Let $\mathcal{D} = \{x_j\}_{j=1}^{N_d}$ denote a dataset of reasoning problems. Given an input $x_j$, a policy $\pi_{\theta}$ samples $M$ independent reasoning completions:
\vspace{-6pt}
\[
\mathcal{Y}_j = \{y_{j,1}, \dots, y_{j,M}\}, \quad y_{j,i} \sim \pi_\theta(\cdot \mid x_j),
\]
where each completion $y_{j,i} = (r_{j,i}, a_{j,i}$) consists of a reasoning trace $r_{j,i}$ and a final answer $a_{j,i}$.



\subsection{Consensus-Based Difficulty Estimation}
For each completion, $y_{j, i}$, we extract its final answer $a_{j, i}$. The empirical agreement ratio for input $x_j$ is defined as:
\begin{equation}
    c_j = \frac{1}{M} \max_{a} \sum_{i=1}^{M} \mathbf{1}[a_{j,i} = a]
\end{equation}

It measures the degree to which independently sampled reasoning trajectories converge to the same solution. High agreement is considered evidence of low epistemic uncertainty and low agreement as an indicator of unresolved ambiguity or reasoning instability. Using a fixed threshold $\rho$, we partition the dataset as:
\begin{equation}
    \mathbb{D}_{\text{easy}} = \{ x_j \mid c_j \geq \rho \}, \quad \mathbb{D}_{\text{hard}} = \{ x_j \mid c_j < \rho \}
\end{equation}

Crucially, this partition is temporary and policy-dependent. Difficulty is recomputed periodically during training, allowing problems to migrate between subsets as the model’s competence evolves.

\subsection{Dynamic Self-Curriculum Training}
Training proceeds in alternating phases of SFT and RL. During SFT, the model is optimized on $\mathbb{D}_{easy}$ using majority-agreed solutions as pseudo-labels. During RL, optimization is restricted to $\mathbb{D}_{hard}$, where the model is encouraged to explore correct yet informative reasoning trajectories under a structured reward. These two phases are interleaved over training, but difficulty reassessment is not performed at every optimization step. Specifically, after a fixed number $K$ of training steps, the current policy $\pi_{\theta}$ is used to regenerate a fresh set of completions for all inputs, agreement ratios are recomputed, and the dataset is re-partitioned
into $\mathbb{D}_{easy}$ and $\mathbb{D}_{hard}$ accordingly. Formally, difficulty reassignment is triggered whenever the training step index $t$ satisfies $t \bmod K = 0$.

This periodic regeneration mechanism induces a closed-loop self-curriculum in which the difficulty of each problem is re-evaluated under the evolving policy. As training progresses, problems that initially exhibit low agreement may transition into the high-consensus regime, while previously easy problems may remain stable. By decoupling difficulty estimation from individual gradient steps and updating it only at fixed intervals, the curriculum remains responsive to policy improvements while avoiding excessive non-stationarity.

\subsection{Supervised Objective}
For inputs $x_j \in \mathbb{D}_{easy}$, we perform supervised fine-tuning using the majority-agreed completion $y_j^\star$.
\begin{equation}
    \mathcal{L}_{\text{SFT}}(\theta) = - \mathbb{E}_{(x_j, y_j^{\star}) \sim D_{\text{easy}}} \left[ \log \pi_{\theta}(y_j^{\star} \mid x_j) \right]
\end{equation}
This phase reinforces reasoning trajectories that are both consensus-consistent and low-variance, consolidating stable inference patterns.


\subsection{Reinforcement Learning Objective}
For each input assigned to the hard subset $\mathbb{D}_{hard}$ we optimize the policy using Group-Relative Policy Optimization (GRPO) over its $N$ sampled completions. Each completion $y_i = (r_i, a_i)$ recieves a reward defined as:
\begin{equation}
\begin{aligned}
R(y_i)
&=
\mathbf{1}[a_i = a_{\text{maj}}(x)]
\cdot
\big(\alpha + \beta\,\mathrm{JSD}_{\mathrm{nov}}(r_i)\big) \\
&\quad \cdot
\big(\varepsilon + (1-\varepsilon)\,g_{\mathrm{rel}}(r_i)\big)
\end{aligned}
\end{equation}
%
which decomposes into correctness gate, population-relative novelty, and relevance-aware gating. This multiplicative structure enforces a strict logical ordering: only only majority-agreed solutions are reinforced; among them, novel reasoning paths are encouraged; and novelty is rewarded only insofar as it remains semantically aligned with the input.

\textbf{Correctness gating.} Let $\{a_i\}_{i=1}^N$ be the set of final answers produced for input $x$. We define the majority answer
\begin{equation}
    a_{\text{maj}}(x) = \mathop{\mathrm{arg\,max}}_{a} \sum_{i=1}^{N} \mathbf{1}[a_i = a]
\end{equation}

and treat it as a pseudo-label induced by self-consistency. The correctness indicator is:
\begin{equation}
    \mathbf{1}[a_i = a_{\text{maj}}(x)]
\end{equation}
implements a consensus-based supervision signal without access to external ground-truth labels. This mechanism provides a conservative yet internally consistent correctness criterion, ensuring that policy updates reinforce only those reasoning trajectories that align with the dominant solution mode under the current policy. An empirical analysis of majority voting is provided in Appendix \ref{app:majorityvotinganalysis}.

\textbf{Population-relative novelty.} To encourage informative exploration in reasoning space, we reward deviation from established population-level solution modes rather than absolute novelty. Let $T_i$ denote the set of reasoning token positions in $r_i$, excluding prompt tokens and the final answer. For each prefix $r_{i, < t}$ the policy induces a next-token distribution $p_\theta (\cdot | x, r_{i, < t})$. We define a majority-conditioned reference distribution by averaging the predictive distributions of reasoning traces whose final answers match the majority mode:
\vspace{-10pt}
\begin{equation}
\bar p_{\mathrm{maj}}(\cdot \mid x, r_{<t}) = 
\frac{1}{|\mathcal{I}_{\mathrm{maj}}(x)|}
\sum_{j \in \mathcal{I}_{\mathrm{maj}}(x)}
p_\theta(\cdot \mid x, r_{j,<t})
\end{equation}
where $\mathcal{I}_{\mathrm{maj}} = \{j : a_j= a_\text{maj}(x)\}$ contains the indices of all sampled reasoning trajectories whose final answers coincide with the majority solution mode for input $x$. The novelty score of a reasoning trace is then defined as the average Jensen–Shannon divergence:
\vspace{-6pt}
\begin{equation}
\mathrm{JSD}_{\mathrm{nov}}(r_i) = 
\frac{1}{|T_i|}
\sum_{t \in T_i}
\mathrm{JS}
\Big(
p_\theta(\cdot \mid x, r_{i,<t})
\Vert
\bar p_{\mathrm{maj}}(\cdot \mid x, r_{<t})
\Big)
\end{equation}
where the Jensen–Shannon divergence is:
\vspace{-6pt}
\begin{equation}
\begin{aligned}
\mathrm{JS}(p \parallel q)
&= \tfrac{1}{2}\,\mathrm{KL}(p \parallel m)
 + \tfrac{1}{2}\,\mathrm{KL}(q \parallel m), \\
m &= \tfrac{1}{2}(p + q)
\end{aligned}
\end{equation}
Unlike unbounded divergences such as KL, which can produce unstable and scale-dependent reward signals due to sensitivity to low-probability events, Jensen–Shannon divergence is bounded and symmetric, providing well-conditioned gradients and a controlled measure of deviation relative to dominant population-level reasoning behavior induced by the current policy.

\textbf{Relevance-aware semantic gating.} Population-relative novelty encourages diversity in reasoning trajectories, but novelty alone does not distinguish between deviations that remain relevant to the input and those arising from semantically unrelated detours. To moderate this effect, we introduce a relevance-aware semantic gate that downweights novelty contributions when intermediate reasoning steps drift away from the input prompt. Each reasoning trace $r_i = \{s_{i,1},\dots, s_{i,n}\}$ is segmented into $r_i = \{s_{i,1},\dots, s_{i,n}\}$ and embedded using a fixed encoder $e(\cdot)$. Relevance is defined by:
\vspace{-8pt}
\begin{equation}
    g_{\mathrm{rel}}(r_i) = 
\frac{1}{n}
\sum_{j=1}^n
\mathrm{clip}
\big(
\cos(e(s_{i,j}), e(x)), 0, 1
\big)
\end{equation}
This gate penalizes reasoning trajectories whose intermediate steps exhibit weak semantic alignment with the input, thereby downweighting novelty that arises from off-topic or weakly related reasoning. The relevance term enters the reward through the affine factor 
\begin{equation}
    \varepsilon + (1-\varepsilon)\cdot g_{\mathrm{rel}}(r_i)
\end{equation}
Here $\varepsilon \in (0,1]$ ensures that rewards are never fully zeroed by low relevance, preventing dead gradients during optimization while still penalizing reasoning trajectories whose semantic content drifts substantially from the input. Overall, the relevance gate $g_\text{rel}$ acts as a soft semantic constraint: it does not enforce correctness or restrict exploration outright, but instead mitigates spurious novelty arising from irrelevant reasoning steps, ensuring that population-relative divergence contributes meaningfully only when deviation remains prompt-aligned.
\vspace{-6pt}
\paragraph{Summary.}
The proposed reward shapes reinforcement learning updates toward reasoning trajectories that are consistent with self-induced consensus, exhibit controlled deviation relative to dominant population behavior, and remain semantically aligned with the input. All reward components are computed exclusively over internal reasoning tokens, with prompt and final answer tokens excluded. Each component plays a complementary role in stabilizing optimization: removing the correctness gate permits reinforcement of inconsistent solution modes, omitting population-relative divergence reduces diversity among reasoning trajectories, and removing relevance-aware gating allows unconstrained novelty to dominate, leading to unstable learning dynamics.

A description of the DiSCTT training procedure is provided in Appendix~\ref{app:pseudocode}.

\section{Results and Discussion}

We next evaluate DiSCTT empirically, focusing on accuracy, stability, computational efficiency, and robustness under distribution shift. Details of the experimental setup are provided in Appendix~\ref{app:experiment-setup}. 

\begin{table*}[!t]
\centering
\caption{
Mean accuracy ($\pm$ standard deviation) across six mathematical and general reasoning benchmarks under test-time adaptation. Results are averaged over independent runs with different random seeds. DiSCTT consistently outperforms the Base model and prior test-time adaptation baselines across model families and scales. Bold denotes the best-performance for each benchmark and model size.
}
\label{tab:main_results_mean_std}
\resizebox{\textwidth}{!}{
\begin{tabular}{llccccccc}
\toprule
\textbf{Model} & \textbf{Method}
& \textbf{AMC}
& \textbf{MATH-500}
& \textbf{AIME-2024}
& \textbf{GPQA}
& \textbf{HotpotQA}
& \textbf{MMLU}
& \textbf{Avg.} \\
\midrule

\rowcolor{lightgray}
\multirow{4}{*}{Qwen-2.5-0.5B-Instruct} & Base        & $12.2_{\pm1.21}$ & $27.6_{\pm0.22}$ & $1.4_{\pm1.89}$ & $14.1_{\pm0.32}$ & $1.2_{\pm0.15}$ & $36.5_{\pm0.13}$ & 15.5  \\
& TTRL        & $20.3_{\pm1.28}$ & $37.2_{\pm0.25}$ & $1.5_{\pm1.66}$ & $16.2_{\pm0.35}$ & $10.3_{\pm0.17}$ & $40.2_{\pm0.16}$ & 10.9 \\
& EVOL-RL     & $23.1_{\pm1.12}$ & $40.0_{\pm0.21}$ & $2.5_{\pm2.10}$ & $18.2_{\pm0.31}$ & $13.7_{\pm0.13}$ & $43.7_{\pm0.14}$ & 23.5 \\
& DiSCTT (Ours) & \textbf{$27.7_{\pm1.11}$} & \textbf{$49.8_{\pm0.23}$} & \textbf{$2.9_{\pm2.33}$} & \textbf{$24.2_{\pm0.33}$} & \textbf{$18.7_{\pm0.14}$} & \textbf{$49.4_{\pm0.12}$} & \textbf{28.7} \\
\midrule

\rowcolor{lightgray}
\multirow{4}{*}{LLaMA-3.2-1B-Instruct} & Base        & $15.4_{\pm1.87}$ & $14.0_{\pm0.24}$ & $0.1_{\pm1.48}$ & $7.1_{\pm0.34}$ & $11.7_{\pm0.16}$ & $14.9_{\pm0.10}$ & 10.5  \\
& TTRL        & $22.5_{\pm1.54}$ & $28.0_{\pm0.21}$ & $1.1_{\pm1.52}$ & $10.6_{\pm0.31}$ & $23.3_{\pm0.18}$ & $29.2_{\pm0.14}$ & 19.1 \\
& EVOL-RL     & $25.4_{\pm1.43}$ & $31.2_{\pm0.23}$ & $1.8_{\pm1.66}$ & $14.1_{\pm0.33}$ & $31.4_{\pm0.15}$ & $36.8_{\pm0.15}$ & 23.4 \\
& DiSCTT (Ours) & \textbf{$26.4_{\pm1.51}$} & \textbf{$39.6_{\pm0.22}$} & \textbf{$2.5_{\pm1.88}$} & \textbf{$23.8_{\pm0.32}$} & \textbf{$40.3_{\pm0.17}$} & \textbf{$47.0_{\pm0.13}$} & \textbf{29.9} \\
\midrule

\rowcolor{lightgray}
\multirow{4}{*}{Qwen-3-1.7B-Base} & Base        & $19.8_{\pm1.72}$ & $47.0_{\pm0.12}$ & $6.6_{\pm2.10}$  & $23.2_{\pm0.22}$ & $25.4_{\pm0.12}$ & $54.2_{\pm0.05}$ & 29.3 \\
& TTRL        & $40.2_{\pm1.20}$ & $55.2_{\pm0.18}$ & $16.3_{\pm1.79}$ & $25.8_{\pm0.28}$ & $49.2_{\pm0.14}$ & $62.7_{\pm0.06}$ & 41.5 \\
& EVOL-RL     & $44.9_{\pm1.97}$ & $53.8_{\pm0.11}$ & $23.3_{\pm1.48}$ & $28.8_{\pm0.21}$ & $56.4_{\pm0.11}$ & $67.3_{\pm0.07}$ & 45.7 \\
& DiSCTT (Ours) & \textbf{$48.8_{\pm1.11}$} & \textbf{$67.0_{\pm0.17}$} & \textbf{$26.6_{\pm1.33}$} & \textbf{$29.3_{\pm0.27}$} & \textbf{$67.1_{\pm0.13}$} & \textbf{$68.4_{\pm0.05}$} & \textbf{51.2} \\
\midrule

\rowcolor{lightgray}
\multirow{4}{*}{LLaMA-3.2-3B-Instruct} & Base        & $12.5_{\pm1.51}$ & $35.7_{\pm0.11}$ & $3.7_{\pm2.33}$ & $13.6_{\pm0.21}$ & $39.9_{\pm0.10}$ & $55.4_{\pm0.07}$ & 26.8 \\
& TTRL        & $27.2_{\pm1.38}$ & $45.4_{\pm0.13}$ & $9.6_{\pm1.63}$ & $16.2_{\pm0.23}$ & $57.3_{\pm0.12}$ & $61.2_{\pm0.08}$ & 36.1 \\
& EVOL-RL     & $32.2_{\pm1.21}$ & $47.2_{\pm0.10}$ & $10.0_{\pm1.52}$ & $20.7_{\pm0.20}$ & $61.7_{\pm0.08}$ & $64.5_{\pm0.07}$ & 39.3 \\
& DiSCTT (Ours) & \textbf{$39.1_{\pm1.12}$} & \textbf{$55.6_{\pm0.16}$} & \textbf{$16.3_{\pm1.89}$} & \textbf{$24.3_{\pm0.26}$} & \textbf{$66.7_{\pm0.09}$} & \textbf{$69.8_{\pm0.06}$} & \textbf{47.8} \\
\midrule

\rowcolor{lightgray}
\multirow{4}{*}{Qwen-3-4B-Base} & Base        & $38.6_{\pm1.20}$ & $51.4_{\pm0.06}$ & $10.0_{\pm1.52}$ & $27.3_{\pm0.11}$ & $39.4_{\pm0.08}$ & $69.5_{\pm0.03}$ & 39.3 \\
& TTRL        & $46.6_{\pm1.38}$ & $60.6_{\pm0.08}$ & $17.1_{\pm1.78}$ & $28.8_{\pm0.13}$ & $61.5_{\pm0.06}$ & $72.4_{\pm0.04}$ & 47.8 \\
& EVOL-RL     & $51.8_{\pm1.12}$ & $64.6_{\pm0.10}$ & $20.4_{\pm1.66}$ & $30.3_{\pm0.15}$ & $66.3_{\pm0.03}$ & $74.5_{\pm0.02}$ & 51.3 \\
& DiSCTT (Ours) & \textbf{$57.0_{\pm0.96}$} & \textbf{$75.2_{\pm0.07}$} & \textbf{$23.7_{\pm1.12}$} & \textbf{$38.4_{\pm0.12}$} & \textbf{$72.9_{\pm0.03}$} & \textbf{$81.3_{\pm0.01}$} & \textbf{58.1} \\
\midrule

\rowcolor{lightgray}
\multirow{4}{*}{Qwen-2.5-7B-Instruct} & Base        & $39.6_{\pm1.21}$ & $58.8_{\pm0.05}$ & $10.7_{\pm1.33}$ & $27.3_{\pm1.09}$ & $51.2_{\pm0.03}$ & $76.2_{\pm0.04}$ & 43.9 \\
& TTRL        & $51.1_{\pm1.12}$ & $74.2_{\pm0.07}$ & $20.3_{\pm1.11}$ & $28.8_{\pm1.11}$ & $62.1_{\pm0.02}$ & $76.4_{\pm0.03}$ & 52.1 \\
& EVOL-RL     & $55.0_{\pm0.96}$ & $73.4_{\pm0.05}$ & $26.3_{\pm1.09}$ & $29.3_{\pm1.09}$ & $66.1_{\pm0.04}$ & $77.9_{\pm0.05}$ & 54.6 \\
& DiSCTT (Ours) & \textbf{$59.5_{\pm0.87}$} & \textbf{$82.2_{\pm0.06}$} & \textbf{$29.6_{\pm0.98}$} & \textbf{$34.9_{\pm0.10}$} & \textbf{$73.7_{\pm0.01}$} & \textbf{$83.3_{\pm0.02}$} & \textbf{60.6} \\

\bottomrule
\end{tabular}
}
\vspace{-10pt}
\end{table*}
Table \ref{tab:main_results_mean_std} lists performance across six challenging mathematical and general reasoning benchmarks. 
Across all evaluated model families and scales, DiSCTT achieves higher mean accuracy than the Base model and strong test-time adaptation baselines, including TTRL and EVOL-RL, and showing consistently lower variance across independent runs. 

The gains are most pronounced on datasets where DiSCTT assigns a larger fraction of updates to supervised fine-tuning (Table \ref{tab:Data-distribution}), which acts as a low-variance pre-adaptation stage that aligns the model with the target distribution prior to reinforcement learning. 
This effect is particularly evident on MATH-500, MMLU, and HotpotQA, where SFT-dominant or balanced SFT/RL splits enable DiSCTT to refine task-specific reasoning priors before exploration. 

Overall, these results show that difficulty-aware bifurcation 
yields a more reliable and scalable test-time adaptation strategy than applying reinforcement learning uniformly across inputs. 
Moreover, DiSCTT maintains strong performance under distribution shift, consistently outperforming uniform test-time training baselines on out-of-distribution reasoning benchmarks. We present ablation studies next.

\subsection{Out-of-Distribution Generalization}
\begin{figure}[ht]
  \vskip 0.2in
  \begin{center}
    \centerline{\includegraphics[width=\columnwidth]{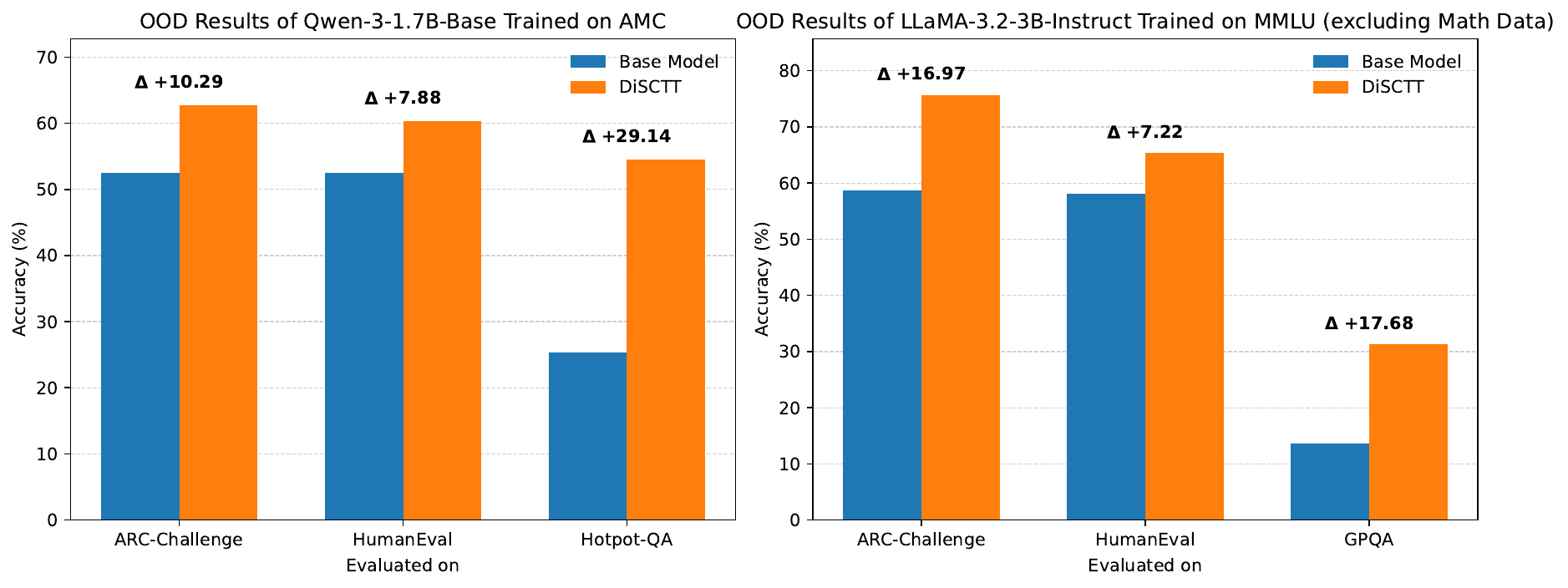}}
    \caption{Out-of-distribution (OOD) reasoning performance under test-time adaptation. DiSCTT consistently outperforms baselines, indicating improved robustness to distribution shift without sacrificing in-distribution performance.}
    \label{OOD-fig}
  \end{center}
  \vspace{-24pt}
\end{figure}

A key risk in test-time adaptation is over-specialization and catastrophic forgetting when optimization is applied to a narrow data source. Figure \ref{OOD-fig} shows that DiSCTT avoids this failure mode while consistently improving out-of-distribution (OOD) performance. Adapting Qwen-3-1.7B-Base on AMC yields large OOD gains (+10.29 ARC-Challenge, +7.88 HumanEval, +29.14 HotpotQA), including code generation, while adapting LLaMA-3.2-3B-Instruct on MMLU (excluding math) achieves similarly strong OOD improvements (+16.97 ARC-Challenge, +17.68 GPQA) with no regressions, demonstrating robust cross-domain generalization. 

These results indicate that DiSCTT preserves general reasoning capability while improving performance in uncertain regimes. This robustness arises from difficulty-aware routing, which avoids unnecessary high-variance updates on high-consensus instances, and from majority-anchored rewards that constrain exploration to knowledge-consistent reasoning regions. Together, these mechanisms limit overfitting to spurious test-time correlations and stabilize adaptation, leading to improved out-of-distribution generalization without sacrificing model stability.

\vspace{-18pt}

\subsection{Training Paradigms Across Difficulty Levels}
\begin{figure*}[!ht]
  \begin{center}
    \centerline{\includegraphics[scale=0.3]{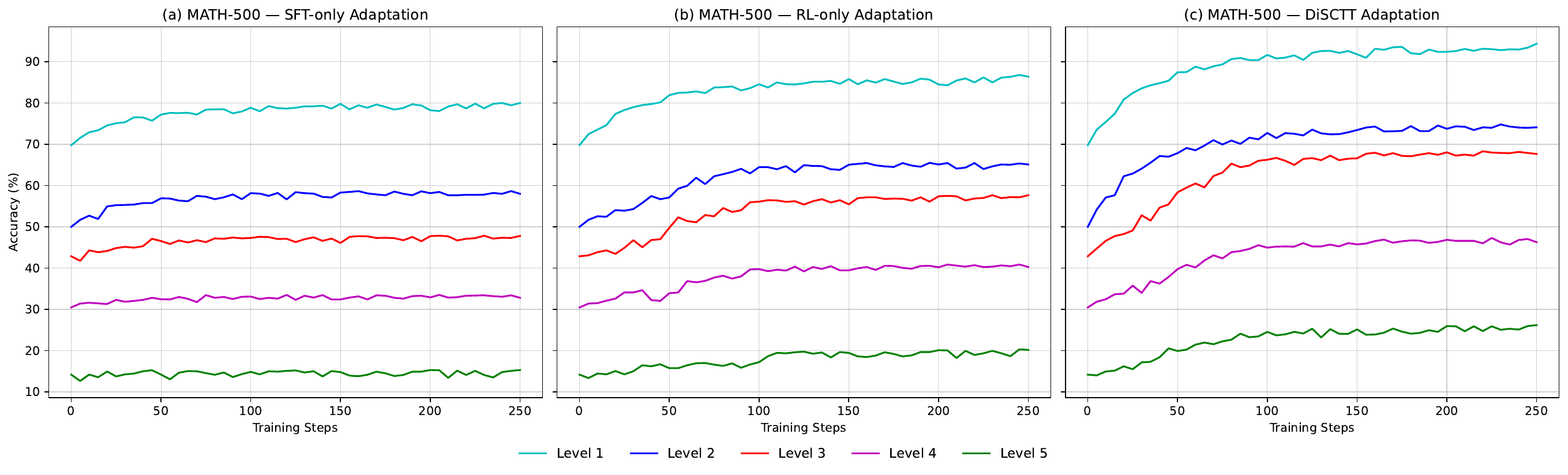}}
    \caption{
    Difficulty-level accuracy dynamics on MATH-500 under different training paradigms. We compare (a) SFT-only, (b) RL-only (GRPO), and (c) DiSCTT across five difficulty levels (L1–L5) over adaptation steps.
    }
    \label{Difficulty-fig}
  \end{center}
 \vspace{-6pt}
\end{figure*}

We analyze the effect of different training paradigms—SFT-only, RL-only (GRPO), and DiSCTT—across difficulty levels on MATH-500, as shown in Figure \ref{Difficulty-fig}. This ablation isolates the role of curriculum routing and hybrid optimization in test-time training.

Figure \ref{Difficulty-fig}(a) shows that SFT-only adaptation is insufficient for test-time training, particularly on harder problems. While SFT steadily improves accuracy on lower-difficulty instances (Levels 1–3) as training progresses, it fails to yield meaningful gains on higher-difficulty levels (Levels 4–5). Moreover, as also reflected by the overall saturation trend in Figure \ref{fig:2}, continued SFT exhibits diminishing returns and can even lead to slight degradation on difficult instances after extended updates, consistent with overfitting to high-consensus patterns rather than improving the model’s underlying reasoning on low-consensus questions. This highlights a key limitation of purely supervised adaptation for challenging reasoning problems.

Figure \ref{Difficulty-fig}(b) presents RL-only (GRPO) adaptation, which consistently outperforms SFT across all difficulty levels, especially on harder problems. However, RL exhibits slower convergence, with delayed improvements on Levels 4 and 5. This behavior indicates that while RL enables exploration, it requires substantial interaction and adaptation time to yield stable reasoning improvements. Even after convergence, the gains on difficult instances remain limited, suggesting that RL alone struggles with early-stage instability and inefficient exploration.

In contrast, Figure \ref{Difficulty-fig}(c) demonstrates that DiSCTT achieves both faster and stronger improvements across all difficulty levels. By first applying SFT on high-consensus instances and dynamically routing low-consensus instances to RL, DiSCTT benefits from a strong initialization before RL begins. This results in a pronounced early performance jump and sustained improvements, particularly on Levels 4 and 5. Compared to SFT-only and RL-only baselines, DiSCTT consistently achieves higher accuracy and better stability, validating the effectiveness of difficulty-aware routing and curriculum-based alternation between supervised and reinforcement learning.

\subsection{Computational Efficiency and Cost-Accuracy Trade-off}
DiSCTT achieves a substantially improved cost–accuracy trade-off over monolithic test-time RL. As shown in Tables \ref{tab:compute_efficiency}(a) and \ref{tab:compute_efficiency}(b), DiSCTT consistently attains higher accuracy while requiring up to 50\% less computation (both FLOPs and wall-clock time) than TTRL across model scales and datasets. For example, on LLaMA-3.2-1B (MMLU), DiSCTT reduces total computation from  $86.44 \times 10^{18}$ to $47.08 \times 10^{18}$ FLOPs while improving accuracy by +17 points, dominating TTRL across evaluated settings. 

These gains stem from DiSCTT’s difficulty-aware routing, which addresses a key inefficiency of prior methods: uniformly applying costly RL updates regardless of instance difficulty. As summarized in Table \ref{tab:Data-distribution}, high-consensus datasets (e.g., MMLU, HotpotQA) naturally favor lightweight SFT updates, whereas intrinsically difficult datasets (e.g., AIME-2024) invoke RL almost exclusively. This adaptive allocation avoids diminishing returns from RL on easy instances and concentrates exploration where policy improvement is most impactful. From a systems perspective, DiSCTT reduces wasted optimization effort and improves learning efficiency per FLOP, with benefits that compound as model size increases. Overall, DiSCTT demonstrates that selective, difficulty-aware adaptation can dominate uniform test-time RL in both accuracy and efficiency, offering a more scalable paradigm for reasoning-centric model adaptation.

\begin{table}[t]
    \centering
    \begin{tabular}{lcc}
        \toprule
        Dataset   & SFT (\%) & RL (\%) \\
        \midrule
        AMC       & 25.0     & 75.0 \\
        MATH-500  & 47.0     & 53.0 \\
        AIME-2024 & 3.3      & 96.7 \\
        GPQA      & 28.8     & 71.2 \\
        HotpotQA  & 58.3     & 41.7 \\
        MMLU      & 67.1     & 32.9 \\
        \bottomrule
    \end{tabular}
    \caption{Average percentage of instances routed to supervised fine-tuning (SFT) and reinforcement learning (RL) across models for each dataset under consensus-based difficulty routing.}
    \label{tab:Data-distribution}
\end{table}

\vspace{0.1pt}
\begin{table*}[t]
\centering
\caption{
\textbf{Compute efficiency comparison between DiSCTT and TTRL.} Tables present (a) total training compute in FLOPs and (b) wall-clock training time. By routing high-consensus instances to SFT and applying RL only to low-consensus instances, DiSCTT substantially reduces compute relative to RL-only TTRL. \emph{Cost Ratio} denotes relative compute cost (DiSCTT/TTRL).
}
\label{tab:compute_efficiency}

\textbf{(a) Training Compute (FLOPs)}
\vspace{2pt}

\begin{tabular}{llccccc}
\toprule
Model & Approach & Dataset & SFT & RL & Total & Cost Ratio\\
\midrule
\multirow{2}{*}{Llama-3.2-1B-Instruct} & DiSCTT & MMLU
& $2.03 \times 10^{16}$ & $47.06 \times 10^{18}$ & $47.08 \times 10^{18}$
& \multirow{2}{*}{0.544 (\textcolor{ForestGreen}{+45.6\%})} \\
& TTRL & MMLU & 0 & $86.44 \times 10^{18}$ & $86.44 \times 10^{18}$ & \\
\midrule
\multirow{2}{*}{Qwen3-4B-Base} & DiSCTT & MATH-500
& $5.77 \times 10^{16}$ & $22.41 \times 10^{18}$ & $22.46 \times 10^{18}$
& \multirow{2}{*}{0.683 (\textcolor{ForestGreen}{+31.7\%})} \\
& TTRL & MATH-500 & 0 & $32.78 \times 10^{18}$ & $32.78 \times 10^{18}$ & \\
\midrule
\multirow{2}{*}{Qwen2.5-7B-Instruct} & DiSCTT & AMC
& $4.45 \times 10^{15}$ & $8.50 \times 10^{18}$ & $8.50 \times 10^{18}$
& \multirow{2}{*}{0.573 (\textcolor{ForestGreen}{+43.7\%})} \\
& TTRL & AMC & 0 & $14.05 \times 10^{18}$ & $14.05 \times 10^{18}$ & \\
\bottomrule
\end{tabular}

\vspace{8pt}

\textbf{(b) Wall-Clock Training Time}
\vspace{2pt}

\begin{tabular}{llccccc}
\toprule
Model & Approach & Dataset & SFT & RL & Total & Cost Ratio \\
\midrule
\multirow{2}{*}{Llama-3.2-1B-Instruct} & DiSCTT & MMLU
& 134\,min & 121\,h\,42\,m & 123\,h\,56\,m
& \multirow{2}{*}{0.513 (\textcolor{ForestGreen}{+48.7\%})} \\
& TTRL & MMLU & 0 & 241\,h\,23\,m & 241\,h\,23\,m & \\
\midrule
\multirow{2}{*}{Qwen3-4B-Base} & DiSCTT & MATH-500
& 21\,min & 51\,h\,48\,m & 52\,h\,9\,m
& \multirow{2}{*}{0.596 (\textcolor{ForestGreen}{+40.4\%})} \\
& TTRL & MATH-500 & 0 & 87\,h\,23\,m & 87\,h\,23\,m & \\
\midrule
\multirow{2}{*}{Qwen2.5-7B-Instruct} & DiSCTT & AMC
& 6\,min & 15\,h\,29\,m & 15\,h\,35\,m
& \multirow{2}{*}{0.521 (\textcolor{ForestGreen}{+47.9\%})} \\
& TTRL & AMC & 0 & 29\,h\,53\,m & 29\,h\,53\,m & \\
\bottomrule
\end{tabular}
\end{table*}

\subsection{Reward Decomposition}
To isolate the contribution of individual reward components, Table \ref{tab:Reward-Decomp} presents a controlled ablation on Qwen-3-4B-Base. Starting from the Base model (51.4\%), adding only the correctness gate improves performance to 62.8\%, confirming that majority-based filtering alone already stabilizes RL updates. Introducing the Population-relative novelty further boosts accuracy to 73.4\%, highlighting the importance of rewarding informative deviations relative to the consensus distribution rather than absolute entropy. Finally, incorporating the consensus-conditioned novelty gate yields the best performance (75.0\%), demonstrating that Relevance-aware semantic gating effectively suppresses spurious diversity while still encouraging structurally distinct reasoning paths.

\begin{table}[H]
    \centering
    \begin{tabular}{lc}
        \toprule
        Dataset   &Accuracy (\%) \\
        \midrule
        - Qwen3-4B-Base                         & 51.4     \\
        - Correctness gate                      & 62.8     \\
        - Population-relative novelty           & 73.4     \\
        - Relevance-aware semantic gating       & 75.0     \\
        \bottomrule
    \end{tabular}
    \caption{Impact of reward term decomposition on model performance for MATH-500 dataset}
    \label{tab:Reward-Decomp}
\end{table}

\section{Conclusion}
We introduced DiSCTT, a test-time adaptation framework for reasoning models that leverages Difficulty-aware self-curriculum learning to improve performance, stability, and efficiency without access to external supervision. By allocating different optimization strategies to inputs based on online estimates of difficulty, DiSCTT enables models to consolidate reliable reasoning behavior while selectively exploring uncertain inference regimes.

Across a diverse set of mathematical and general reasoning benchmarks, DiSCTT achieves consistent improvements over strong test-time adaptation baselines, exhibiting lower variance and reduced optimization cost. These gains arise from how adaptation effort is structured rather than from increased optimization, underscoring the importance of difficulty-aware allocation in test-time learning.

More broadly, our results suggest that explicitly accounting for instance-level uncertainty and heterogeneity is central to effective test-time adaptation in reasoning systems. Difficulty-aware self-curriculum learning offers a principled alternative to uniform test-time training and may serve as a reusable design pattern for adaptive inference in LLMs. Future work may explore extensions to other modalities, tighter uncertainty estimates, and theoretical characterizations of curriculum-driven test-time learning dynamics.

\section*{Accessibility}
We aim to make this work accessible to a broad audience, including readers with disabilities or sensory and neurological differences.

\section*{Software and Data}

Code and experimental data to be released upon publication.



\nocite{karras2022elucidating}
\section*{Impact Statement}
This work advances test-time adaptation for reasoning models by improving the efficiency and stability of inference-time learning without requiring additional supervision. By reallocating adaptation effort based on instance difficulty, the proposed approach reduces computational cost while improving performance on complex reasoning tasks. Improved efficiency and robustness may broaden access to adaptive reasoning systems and support applications in education, scientific analysis, and decision support. As with all adaptive language models, these capabilities may also be misused, including for generating misleading or overconfident reasoning. We view the proposed method as a neutral algorithmic contribution and emphasize the importance of responsible deployment and appropriate safeguards.

\nocite{langley00}
\bibliography{example_paper}

@inproceedings{langley00,
 author    = {P. Langley},
 title     = {Crafting Papers on Machine Learning},
 year      = {2000},
 pages     = {1207--1216},
 editor    = {Pat Langley},
 booktitle     = {Proceedings of the 17th International Conference
              on Machine Learning (ICML 2000)},
 address   = {Stanford, CA},
 publisher = {Morgan Kaufmann}
}

@article{wei2022chain,
  title={Chain-of-thought prompting elicits reasoning in large language models},
  author={Wei, Jason and Wang, Xuezhi and Schuurmans, Dale and Bosma, Maarten and Xia, Fei and Chi, Ed and Le, Quoc V and Zhou, Denny and others},
  journal={Advances in neural information processing systems},
  volume={35},
  pages={24824--24837},
  year={2022}
}

@article{ouyang2022training,
  title={Training language models to follow instructions with human feedback},
  author={Ouyang, Long and Wu, Jeffrey and Jiang, Xu and Almeida, Diogo and Wainwright, Carroll and Mishkin, Pamela and Zhang, Chong and Agarwal, Sandhini and Slama, Katarina and Ray, Alex and others},
  journal={Advances in neural information processing systems},
  volume={35},
  pages={27730--27744},
  year={2022}
}

@inproceedings{sun2020test,
  title={Test-time training with self-supervision for generalization under distribution shifts},
  author={Sun, Yu and Wang, Xiaolong and Liu, Zhuang and Miller, John and Efros, Alexei and Hardt, Moritz},
  booktitle={International conference on machine learning},
  pages={9229--9248},
  year={2020},
  organization={PMLR}
}

@article{zuo2025ttrl,
  title={Ttrl: Test-time reinforcement learning},
  author={Zuo, Yuxin and Zhang, Kaiyan and Sheng, Li and Qu, Shang and Cui, Ganqu and Zhu, Xuekai and Li, Haozhan and Zhang, Yuchen and Long, Xinwei and Hua, Ermo and others},
  journal={arXiv preprint arXiv:2504.16084},
  year={2025}
}

@article{wang2022self,
  title={Self-consistency improves chain of thought reasoning in language models},
  author={Wang, Xuezhi and Wei, Jason and Schuurmans, Dale and Le, Quoc and Chi, Ed and Narang, Sharan and Chowdhery, Aakanksha and Zhou, Denny},
  journal={arXiv preprint arXiv:2203.11171},
  year={2022}
}

@article{moradi2025continuous,
  title={Continuous Self-Improvement of Large Language Models by Test-time Training with Verifier-Driven Sample Selection},
  author={Moradi, Mohammad Mahdi and Amer, Hossam and Mudur, Sudhir and Zhang, Weiwei and Liu, Yang and Ahmed, Walid},
  journal={arXiv preprint arXiv:2505.19475},
  year={2025}
}

@article{zhou2025evolving,
  title={Evolving language models without labels: Majority drives selection, novelty promotes variation},
  author={Zhou, Yujun and Liang, Zhenwen and Liu, Haolin and Yu, Wenhao and Panaganti, Kishan and Song, Linfeng and Yu, Dian and Zhang, Xiangliang and Mi, Haitao and Yu, Dong},
  journal={arXiv preprint arXiv:2509.15194},
  year={2025}
}

@article{christiano2017deep,
  title={Deep reinforcement learning from human preferences},
  author={Christiano, Paul F and Leike, Jan and Brown, Tom and Martic, Miljan and Legg, Shane and Amodei, Dario},
  journal={Advances in neural information processing systems},
  volume={30},
  year={2017}
}

@article{shao2024deepseekmath,
  title={Deepseekmath: Pushing the limits of mathematical reasoning in open language models},
  author={Shao, Zhihong and Wang, Peiyi and Zhu, Qihao and Xu, Runxin and Song, Junxiao and Bi, Xiao and Zhang, Haowei and Zhang, Mingchuan and Li, YK and Wu, Yang and others},
  journal={arXiv preprint arXiv:2402.03300},
  year={2024}
}

@inproceedings{goyal2022revisiting,
  title={Revisiting Test-Time Adaptation under Distribution Shifts},
  author={Goyal, Anirudh and others},
  booktitle={ICML},
  year={2022}
}

@article{zhang2023testtime,
  title={Test-Time Adaptation for Language Models},
  author={Zhang, Ziqi and others},
  journal={arXiv preprint arXiv:2306.05644},
  year={2023}
}

@inproceedings{wei2022cot,
  title={Chain-of-Thought Prompting Elicits Reasoning in Large Language Models},
  author={Wei, Jason and Wang, Xuezhi and Schuurmans, Dale and Bosma, Maarten and Ichter, Brian and Xia, Fei and Chi, Ed H. and Le, Quoc V. and Zhou, Denny},
  booktitle={Advances in Neural Information Processing Systems (NeurIPS)},
  year={2022}
}

@inproceedings{geng2024survey,
  title={A survey of confidence estimation and calibration in large language models},
  author={Geng, Jiahui and Cai, Fengyu and Wang, Yuxia and Koeppl, Heinz and Nakov, Preslav and Gurevych, Iryna},
  booktitle={Proceedings of the 2024 Conference of the North American Chapter of the Association for Computational Linguistics: Human Language Technologies (Volume 1: Long Papers)},
  pages={6577--6595},
  year={2024}
}
\bibliographystyle{icml2026}

\newpage
\appendix
\onecolumn
\section{Algorithm Pseudocode}
\label{app:pseudocode}
Algorithm \ref{alg:discct} presents the full pseudocode of Difficulty-Aware Self-Curriculum Test-Time Adaptation (DiSCTT). The algorithm maintains a dynamic partition of test inputs based on an online estimate of instance difficulty derived from consensus among multiple sampled reasoning completions. At periodic intervals, the current policy generates $M$ independent reasoning traces for each input and computes a consensus score measuring agreement among final answers. Inputs with consensus above a threshold $\rho$ are treated as high-consensus (easy) instances, while the remainder are classified as low-consensus (hard). DiSCTT then applies supervised fine-tuning to consolidate majority-agreed solutions on easy instances, and reinforcement learning via GRPO on hard instances using a majority-anchored novelty reward. The difficulty partition is reused between re-routing steps and recomputed periodically, enabling a self-evolving curriculum in which instances migrate between easy and hard as the policy adapts. This procedure enables fully label-free, test-time self-improvement by explicitly allocating optimization objectives according to epistemic uncertainty estimated from the model’s own generations.

\begin{algorithm}[H]
\caption{Difficulty-Aware Self-Curriculum Test Time Adaptation (DiSCTT)}
\label{alg:discct}
\begin{algorithmic}[1]
\REQUIRE Dataset $\mathcal{D}$, policy $\pi_{\theta}$, samples $M$, update interval $K$, consensus threshold $\rho$
\STATE Initialize $\theta \leftarrow \theta_0$
\FOR{training step $t=1,2,\dots$}
    \IF{$t \bmod K = 0$}
    \FOR{each input $x_j \in \mathcal{D}$} 
        \STATE Sample $M$ reasoning completions\\ $\{y_{j,i}=(r_{j,i},a_{j,i})\}_{i=1}^M \sim \pi_\theta(\cdot|x_j)$
        \STATE Compute consesus score\\ $c_j \leftarrow \frac{1}{M} \max_a \sum_{i=1}^M \mathbb{I}[a_{j,i}=a]$
        \ENDFOR
        \STATE $\mathcal{D}_{\mathrm{easy}}=\{x_j:c_j\ge\rho\},\;
               \mathcal{D}_{\mathrm{hard}}=\mathcal{D}\setminus\mathcal{D}_{\mathrm{easy}}$
    \ENDIF
    \STATE Use the most recent difficulty partition for intermediate steps
    \STATE $y_j^\star \leftarrow \text{majority-agreed completion for } x_j$
    \STATE \textbf{SFT:} $\;\min_\theta\;-\log\pi_\theta(y_j^\star|x_j),\; x_j\in\mathcal{D}_{\mathrm{easy}}$
    \STATE \textbf{RL (GRPO):} $\;\max_\theta\;\mathbb{E}[R(y_{j,i})],\; x_j\in\mathcal{D}_{\mathrm{hard}}$
\ENDFOR
\STATE Return updated $\theta$
\end{algorithmic}
\end{algorithm}

\section{Experimental Setup} 
\label{app:experiment-setup}
We evaluate DiSCTT on six diverse reasoning benchmarks spanning mathematical, multi-hop, and knowledge-intensive reasoning. Specifically, we consider four mathematical reasoning datasets (AMC, MATH-500, AIME-2024, GPQA), one multi-hop question answering benchmark (HotpotQA), and one broad knowledge and reasoning benchmark (MMLU). This selection enables a comprehensive assessment of the proposed method across tasks with varying levels of difficulty, structure, and reasoning requirements.

To demonstrate the generality and scalability of DiSCTT, we conduct experiments across six language models of varying sizes, including both Base and Instruct-tuned variants. This setting allows us to evaluate the effectiveness of DiSCTT across different initialization regimes and model capacities.

All experiments are performed on a single machine equipped with 8 NVIDIA RTX A6000 GPU (50 GB memory). Unless stated otherwise, we use a consistent set of training hyperparameters across all experiments. In the consensus-based difficulty estimation stage, for each input prompt 
$x$, we generate $M=8$ independent completions. The consensus threshold $\rho$ used to partition data between supervised fine-tuning (SFT) and reinforcement learning (RL) is set to 
$45\%$. Difficulty estimation is recomputed periodically every curriculum cycle, where each cycle consists of 2 SFT epochs followed by 10 RL epochs, after which the consensus-based partitioning is updated. The training sequence length is fixed to 2048 tokens throughout all experiments.

LoRA adaptation is applied in both the SFT and RL phases, with rank $\text{lora}_r = 32$ and scaling factor $\text{lora}_{alpha}=64$. For supervised fine-tuning, the batch size is set to 8, the learning rate is $1e-5$, and the total number of SFT epochs is 8, corresponding to 2 epochs per curriculum round. For reinforcement learning, we perform a total of 40 epochs, with 10 epochs per curriculum round. The RL batch size is 8, with 32 sampled completions per input, a learning rate of $1e-6$, and a sampling temperature of 0.9. The reward function hyperparameters are fixed to $\alpha = 1$, $\beta=1$, and $\epsilon=0.2$ across all experiments.

For the relevance-aware semantic gate $g_{\text{rel}}$, we compute sentence embeddings using the pretrained BAAI/bge-base-en-v1.5 model, which provides a lightweight and robust measure of semantic alignment between reasoning steps and the input prompt.

\section{Additional Experimental Results}
\subsection{Analysis of the Majority Voting as a verifier}
\label{app:majorityvotinganalysis}
Table \ref{tab:MV_Sig} reports the fraction of majority-selected answers that match the ground-truth solution across multiple reasoning benchmarks. For each input, we generate several independent completions and select the majority-agreed answer, using consensus as a proxy correctness signal. Across datasets, majority voting demonstrates an acceptable level of agreement with ground truth, with performance varying by task difficulty and domain. In particular, as shown in Table \ref{tab:MV_Sig}, AIME-2024, which is substantially more challenging than the other benchmarks, exhibits lower agreement between majority-selected answers and ground truth, reflecting the increased uncertainty and diversity of plausible reasoning paths in harder problem regimes.

Compared to verifier-based approaches, majority voting offers practical advantages in the test-time setting. It does not require training or maintaining an external verifier, avoiding additional supervision and potential domain mismatch. Moreover, because the signal is derived directly from the model’s own generations, it naturally adapts as the policy evolves during test-time training. These properties make majority voting a simple and scalable alternative for providing an online epistemic signal within DiSCTT.

\begin{table}[th]
    \centering
    \begin{tabular}{lcccccc}
        \toprule
        Model             & MATH-500 & AIME-2024 & ACM    & GPQA     & HotpotQA & MMLU \\
        \midrule
        Qwen3-1.7B-Base & 76.2\%   & 50.0\%    & 80.0\% &  68.3\%  & 80.1\%   & 70.6\% \\
        \bottomrule
    \end{tabular}
    \caption{Percentage of instances where the majority-selected answer matches the ground truth across reasoning benchmarks}
    \label{tab:MV_Sig}
\end{table}

\end{document}